\documentclass[conference]{IEEEtran}
\usepackage[utf8]{inputenc}
\usepackage[T1]{fontenc}
\usepackage[english]{babel}
\usepackage{graphicx}
\graphicspath{{figures/}}
\usepackage{refstyle}
\usepackage{textcomp}
\usepackage[ruled,vlined,linesnumbered]{algorithm2e}
\usepackage{amsmath}
\usepackage{amsthm}
\usepackage{amssymb}
\usepackage{microtype}
\usepackage{hyperref}

\makeatletter

\usepackage{balance}
\usepackage{csquotes}
\usepackage{subcaption}
\usepackage{siunitx}
\newcommand{\Sec}[1]{Sec.~\ref{#1}}
\newcommand{\Fig}[1]{Fig.~\ref{#1}}

\makeatother
\usepackage[style=ieee, doi=false, isbn=false,backend=bibtex]{biblatex}

\begin{document}

\title{iviz: A ROS Visualization App for Mobile Devices}

\author{\IEEEauthorblockN{\textbf{Antonio~Zea} and \textbf{Uwe~D.~Hanebeck}}
\IEEEauthorblockA{Intelligent Sensor-Actuator-Systems Laboratory (ISAS)\\
Institute for Anthropomatics and Robotics\\
Karlsruhe Institute of Technology (KIT), Germany}
antonio.zea@kit.edu, uwe.hanebeck@kit.edu}
\IEEEpeerreviewmaketitle
\maketitle

\begin{abstract}
In this work, we introduce \emph{iviz}, a mobile application for visualizing ROS data.
In the last few years, the popularity of ROS has grown enormously, making it the standard platform for open source robotic programming.
A key reason for this success is the availability of polished, general-purpose modules for many tasks, such as localization, mapping, path planning, and quite importantly, data visualization.
However, the availability of the latter is generally restricted to PCs with the Linux operating system.
Thus, users that want to see what is happening in the system with a smartphone or a tablet are stuck with solutions such as screen mirroring or using web browser versions of \emph{rviz}, which are difficult to interact with from a mobile interface.
More importantly, this makes newer visualization modalities such as Augmented Reality impossible.
Our application iviz, based on the Unity engine, addresses these issues by providing a visualization platform designed from scratch to be usable in mobile platforms, such as iOS, Android, and UWP, and including native support for Augmented Reality for all three platforms.
If desired, it can also be used in a PC with Linux, Windows, or macOS without any changes.
\end{abstract}


\section{Introduction}\label{sec:introduction}
Since its inception in 2007, the Robot Operating System (ROS) \cite{quigley2009ros} has been gaining ground as the premiere middleware platform for robot programming.
Nowadays, ROS can be seen in every field of robotic, from tiny vacuum cleaners \cite{araujo2015integrating} to lawn tractor \cite{lleras2016development}, from embedded applications \cite {bouchier2013embedded} to operations in the cloud \cite{kehoe2015survey}, from underwater vehicles \cite{demarco2011implementation} to deployments in outer space \cite{badger2016ros}, and even in the decontamination of hazardous environments \cite{petereit2019robdekon}.
The success of ROS can be attributed to several factors, including an open source environment, an enthusiastic community of programmers, and especially in the last years, the support of large robot manufacturers.

Simultaneously, the last few years have seen a renewed interest in exploring new applications and paradigms in robot programming, especially in the context of Industry 4.0 \cite{lu2017industry}.
Topics of interest include how to increase operational efficiency, boost productivity, and achieve a higher level of automatization in industrial settings, ranging from product design to manufacturing pipelines.
Important ideas in this context are communication and interoperability, i.e., the data needs to be accessible exactly where needed, by whoever needs it, at any time.
In robotics, information produced by a robot is usually bound to a position around it, and being able to visualize this data next to the place where it was generated can provide precious contextual information.
This motivates an emphasis on Augmented Reality (AR) and Virtual Reality (VR) technologies, capable of displaying arbitrary three-dimensional information in any point in space.
In general, these modalities of data visualization and interaction are more intuitive and easier to learn than a monitor setup, and have been shown to reduce cognitive load \cite{lee2014learning,baumeister2017cognitive}.

There are multiple projects in literature combining ROS with VR \cite{roldan2019multi,whitney2018ros} and AR \cite{peppoloni2015immersive,lee2018implementation}, especially in an industrial setting \cite{sita2017ros}.
While AR in robotics is not exactly new, and VR even less, they have seen an explosion in growth in the last five to seven years following the appearance of affordable, off-the-shelf devices with accurate and robust user tracking, such as the HTC Vive for VR and the Microsoft Hololens for AR.
And while the Hololens is still a bit pricey at around 4000\texteuro, more affordable alternatives exist in the form of Apple's ARKit and Google's ARCore, which can turn everyday smartphones and tablets into AR presentation devices.

\begin{figure}[t]
    \centering
    \includegraphics[width=0.475\textwidth]{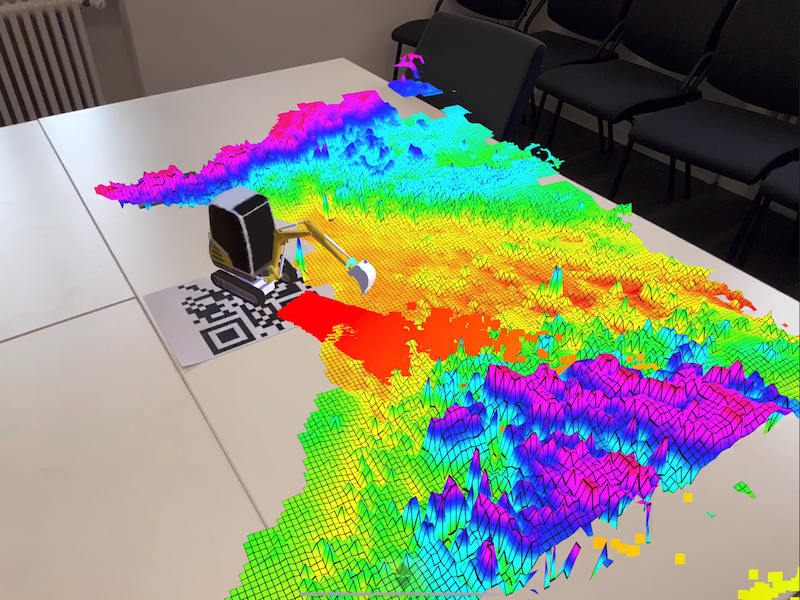}
    \caption{AR visualization of a terrain captured by a robotic excavator from Fraunhofer IOSB \cite{emter2017algorithm}.}
    \label{fig:face}
\end{figure}

While mobile devices provide new and interesting ways to present and interact with information, they do not intersect with the platforms that ROS has traditionally supported, i.e., PCs with Linux using C++ or Python.
And while desktop operating systems such as Windows and macOS have some degree of official support, there are still some unresolved issues which cause most users of these systems to prefer Linux-based virtual machines (or Docker).
Mobile operating systems, however, have received little to no attention --- a notable exception being \cite{rvizforandroid} from 2012. 
This is somewhat justified by the fact that, up until rather recently, most mobile devices were simply not performant enough to do any sort of complex data visualization or data processing. 
Thus, it was difficult to see implementing fully-featured ROS libraries as being worth the required effort.

Workarounds to these issues have centered on intermediary nodes that run on the PC with the ROS installation (i.e., the ROS hub).
On the one hand, a web server can be used to pre-process the data, which can then be presented from the web browser of the device \cite{hilton2014lightweight, webviz}.
While this approach is the simplest way to display ROS data in a mobile device, it also makes AR visualization unfeasible.
On the other hand, if data processing on the device is preferred, the Rosbridge Suite \cite{crick2017rosbridge} can be used instead.
This method employs a `translator' on the hub which transforms ROS messages to a JSON format, and then transmits them to the device using a websocket interface, making the suite accessible from any programming language in any platform.
This, combined with game engines such as Unity \cite{engine2008unity} and libraries such as ROS\# \cite{bischoff2018ros}, has already led to a wide variety of mobile robotic applications in AR such as \cite{kastner2019augmented,muhammad2019creating,manring2020augmented,guhl2017concept}.

\begin{figure*}
    \centering
    \includegraphics[width=0.8\textwidth]{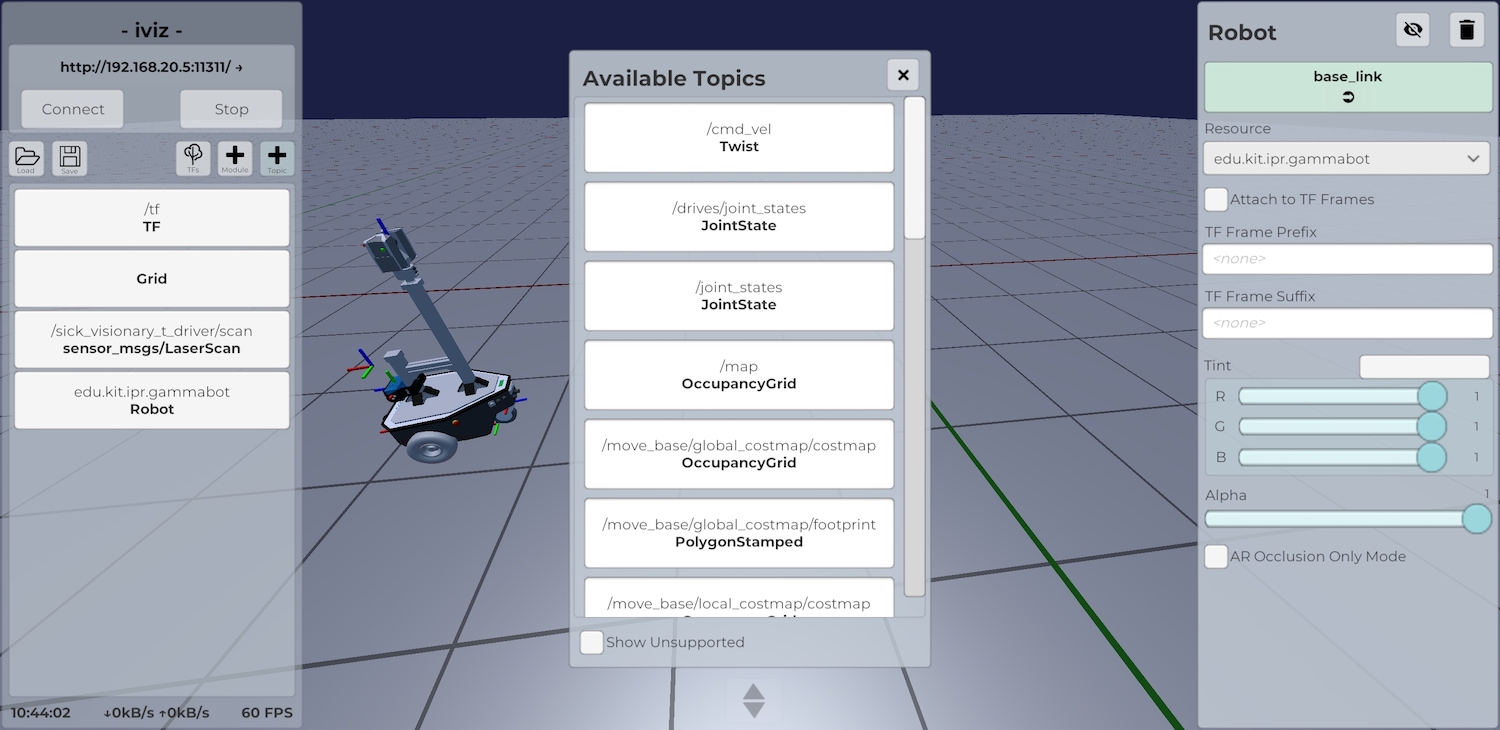}
    \caption{Example screenshot of the iviz GUI with displays (left), available topic (middle), and display settings (right), taken on a Samsung Galaxy S9. The concept is to have `plug-and-play' access to a ROS system: connect, tap topics, and visualize.}
    \label{fig:example_gui}
\end{figure*}

In this paper, we introduce \textbf{iviz}, a new platform for ROS data visualization in mobile devices, in particular Android/iOS smartphones and tables, and the Microsoft Hololens.
We have the following three goals for our platform:
\begin{itemize}
    \item we aim to visualize a wide variety of ROS data, from point clouds to interactive markers,
    \item with fast update rate and (relatively) low latency,
    \item with direct support for AR (as in \Fig{fig:face}),
    \item as self-contained as possible,
    \item and requiring no changes in the existing ROS system.
\end{itemize}
Our application is based on the Unity engine and written in C\#, which allows us to target many operating systems and devices with no changes in the code, and without having to worry about the target OS.
It is fully open source and released under the MIT license.
While we chose a similar name, iviz is not intended to be a replacement for rviz \cite{hershberger2019rviz}, the standard visualization program for ROS.
Instead, our aim is for a mobile app where a user can pull out a tablet or smartphone at any moment, enter a ROS network, and with a couple of taps be able to see what is going on with a robot (\Fig{fig:example_gui}), ideally with AR if desired.
The source code can be found at
\mbox{\underline{\url{https://github.com/KIT-ISAS/iviz/tree/devel}}}, and requires no installation other than cloning the repository.


This paper is structured as follows.
First, we describe the challenges in implementing a multi-platform ROS node in Unity in \Sec{sec:ros_mobile}.
Then, we present the external libraries and auxiliary applications required by iviz in \Sec{sec:external_deps}.
After that, the visualization app and its modules is introduced in \Sec{sec:internal_comps}.
This is followed by a description of our challenges, future work, and a brief conclusion in \Sec{sec:challenges}.

\section{ROS on a Mobile Device}\label{sec:ros_mobile}

ROS is a quasi-operating system with its own drivers, modules, and utilities.
Fortunately, we do not need to rewrite them into our mobile app in order to benefit from them, as ROS provides interfaces that allow their entire functionality to be accessed remotely from the network.
Thus, in order to `talk' to the framework, all we need is to implement its network layer.

In essence, a ROS network is based on decentralized peer-to-peer communication between `nodes', which produce and consume information, and a central `master' whose only task is name resolution and storing parameter values.
The availability of data streams is represented by `topics', and the data itself is discretized in packets called `messages'.
Thus, when a node wants to publish information, it advertises a topic in the master.
Then, a subscriber that is interested in this topic first contacts the master to inquire the IP address of the publisher, and then the publisher directly to request a communication channel.
If a topic has multiple publishers, the subscriber will need to connect to all of them.
The concept of publisher-subscriber has been extended to allow for more fine-grained bidirectional communication in the form of `services' and `actions'.
However, their underlying implementation is essentially the same.

Communicating with this network from a smartphone or a Hololens carries three main challenges.
First, it is necessary to implement the ROS message serialization mechanism in a way that can be used easily from Unity and C\#.
This mechanism also needs to be self-contained, i.e., it should not require files outside of the mobile device.
Second, we need an implementation of the ROS API that handles the connections to the peer nodes.
Third, and finally, a proper visualization app requires access to assets such as robot meshes and textures stored outside of the device.

Addressing these three issues is not straightforward, which is a key reason for why Unity and mobile developers have traditionally chosen to use Rosbridge instead.
However, the used JSON serialization carries two important drawbacks.
On the one hand, the string representation increases the message size dramatically, and even byte arrays encoded in base-64 have an overhead of at least 33\%.
Using BSON alleviates this issue, but the size increase is still significant, among others due to the lack of 32-bit floats and the fact that field names are always transmitted.
On the other hand, commonly used JSON libraries such as \cite{newton2019json,utf8json} speed up string parsing by injecting dynamically generated code into the program, tailored specifically to the expected message structures.
Nowadays, this procedure is disallowed in many mobile operating systems (particularly iOS and UWP), as it has been abused by malware creators in the past.
For this reason, only library variants that are compiled ahead-of-time (AOT) can be used in these devices, which slows down parsing performance significantly and drives up CPU usage unnecessarily.
Another issue with Rosbridge has to do with the network structure it imposes: all messages have to go through the Rosbridge node and then to its destination.
This can introduce a significant bottleneck, especially when streaming high-bandwidth sensor data such as pointclouds and RGB images, leading to dropped packets and latency in other topics too.
It should be pointed out that the new generation of the ROS platform, called `ROS 2', aims to solve many of these problems.
However, given that it is not backwards-compatible, most of the existing code will still require ROS 1 for the foreseeable future.

As we aim for high performance in mobile devices with arbitrary sensor data, a native implementation of the ROS API should be preferred, such as \cite{zoetrope2011}.
Unfortunately, the authors could not find libraries that do not contain dependencies such as XML-RPC.NET \cite{cook2008xml}, which also employ dynamically generated code, making them unportable to mobile.
%
%
For this reason, we found it necessary to implement much of this functionality ourselves and package it in external modules that can also be used by other applications. 
This code, together with the internal workings of our visualizer, will be described in the next sections.

\section{Iviz: External Components}\label{sec:external_deps}
We will now introduce external components tasked with addressing the issues from \Sec{sec:ros_mobile}: message code generation, a ROS API implementation, and a service for loading remote assets.

First, the \textbf{iviz\_msg\_gen} module arose from the need of fast, efficient serialization and deserialization of ROS messages without external dependencies, or at least without dependencies that cannot be easily packed into a mobile device.
A ROS message is a structure that contains a list of fields, which can be either primitives (such as integers, floats, or strings), arrays, or other messages.
An approach used by official libraries, such as \emph{libcpp}, is to convert these files into native C++ code that contains all the relevant information about the message (versioning, message size, serialization instructions), which is then treated as an additional source file in the user's project.
For our library, we use the same idea, and generate a self-contained C\# file for each message, incorporating MD5 hashes and dependency declarations as string constants.
This automatically generated code is then written to the \textbf{iviz\_msgs} project, and compiled into a stand-alone library.

Second, an implementation of the ROS API is provided by the module  \textbf{iviz\_roslib}, which can be used by any C\# application to talk to ROS nodes.
The interface is similar to that provided by ROS\#, making portability from existing projects straightforward.
Implemented features include publishers, subscribers, services, and accessing the parameter server.
Note that our library only implements the TCPROS transport protocol, and support for UDP is left for future work.

Finally, the third optional component aims to solve the issue of external assets, such as 3D modules or textures, generally stored in the ROS hub.
As mentioned before, these assets are needed by the mobile device to display entities such as robots, markers, or the world in the background.
Simply copying those files to the mobile device is not straightforward.
On the one hand, unlike photos or videos, there is (in the subjective view of the authors) no convenient way to copy arbitrary files from a PC to arbitrary folders in a mobile device, and searching individual COLLADA files quickly becomes bothersome.
On the other hand, we would still need an external library for opening those 3D models, as Unity only provides this functionality within the editor.
To address this, we introduce \textbf{iviz\_loader\_service}, a service node that runs on the PC with the assets.
Then, whenever an external model or texture is needed by the mobile visualization, the service is called by the application, which in turn locates the asset, opens it with the Assimp library \cite{schulze2012open}, applies pre-processing steps, and transmits its contents back to the device.
This service functionality is invoked transparently without requiring input from the user, and the processed assets are cached locally in the device for future use.
A slight drawback of this approach is the need to have a service node running in the ROS system, which can introduce a security risk, especially if some of the models contain sensitive information or are protected by NDAs.

\section{Iviz: Internal Modules}\label{sec:internal_comps}
A key focus in the design was portability: the same modules can be used in Windows, Linux, macOS, but also iOS, Android and UWP, with minimal or no changes.
For platforms that do not support C\# directly, Unity provides an utility called `IL2CPP' which transforms C\# code to C++. 
Furthermore, while the iviz application was implemented as a Unity project, the code should be easily portable to other game engines that support C\#, such as the open source Godot \cite{linietsky2019godot}.

\begin{figure}[ht]
    \centering
    \includegraphics[width=0.475\textwidth]{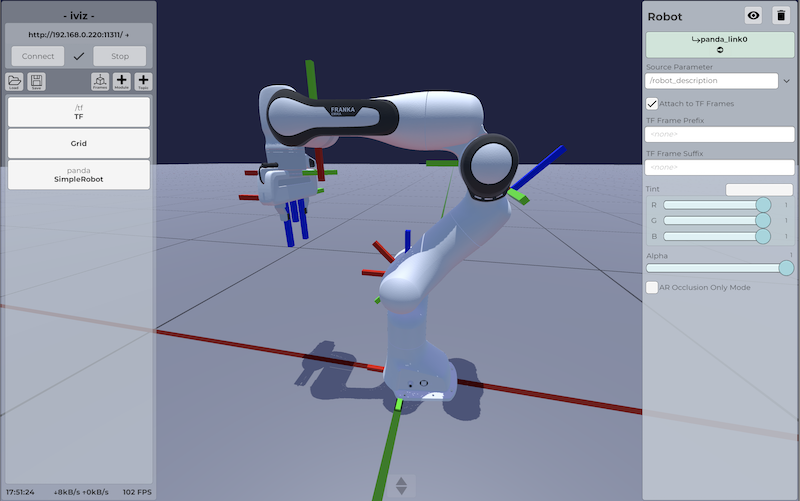}
    \caption{A Panda manipulator from Franka Emika, visualized on a macOS laptop. The 3D models were downloaded on the fly from a Linux hub.}
    \label{fig:robot}
\end{figure}

Similar to rviz, the visualization in iviz is based around the idea of `displays', i.e., reusable and recyclable code modules tasked with rendering entities such as lines, point clouds, duplicated meshes, etc.
For instance, a module in charge of displaying a \emph{PointCloud2} message can also be reused to display a \emph{Point List} marker.
Similarly, the instantiation method that allows for displaying point clouds as cubes can be applied to a \emph{Cube List} marker, or the axes in a \emph{Path} message, and so on.
This approach also brings two benefits in Unity.
On the one hand, Unity detects when multiple objects in the scene are using the same materials, meshes, or textures, and fuses them together into a single draw call, making rendering faster and cheaper.
On the other hand, instantiation of new Unity objects can be rather expensive. 
We can avoid these costs by employing a resource pool that recycles discarded components, such as line renderers and meshes, instead of having to destroy re-create them constantly.
For example, if a frame axis made of cubes is no longer needed, the cube components can be reused as the links of a robot, and so on.
This approach allows for highly dynamic scenes, especially on mobile devices, without instantiation becoming a bottleneck.

The implemented displays and modules include
\begin{itemize}
    \item pointclouds,
    \item laser scans,
    \item (interactive) markers,
    \item joystick publishers,
    \item occupancy grid,
    \item grid maps,
\end{itemize}
among others.
In addition to that, two modules stand out for discussions are the robot module and the AR module, explained in the following sections.

\subsection{Robot Module}\label{sec:robot_module}
Robots necessarily play a central component of any ROS visualization platform.
However, similarly to 3D assets, their visualization is made difficult by the fact that their specification files are outside of the device.
Fortunately, ROS applications automatically upload the specification of the current robot to the parameter server, which can then be easily read by our application.
Nonetheless, unless the visuals consist exclusively of cubes and cylinders, we still need a way to access the CAD models.
To address this, the robot module is programmed to contact the iviz\_loader\_service node automatically (if available) to retrieve these missing resources (\Fig{fig:robot}).
As the specification also contains information about the joints and their positions, iviz also supports setting robot configurations directly, for example, by listening to a \emph{JointState} topic.



\begin{figure*}
    \centering
    \includegraphics[width=0.7\textwidth]{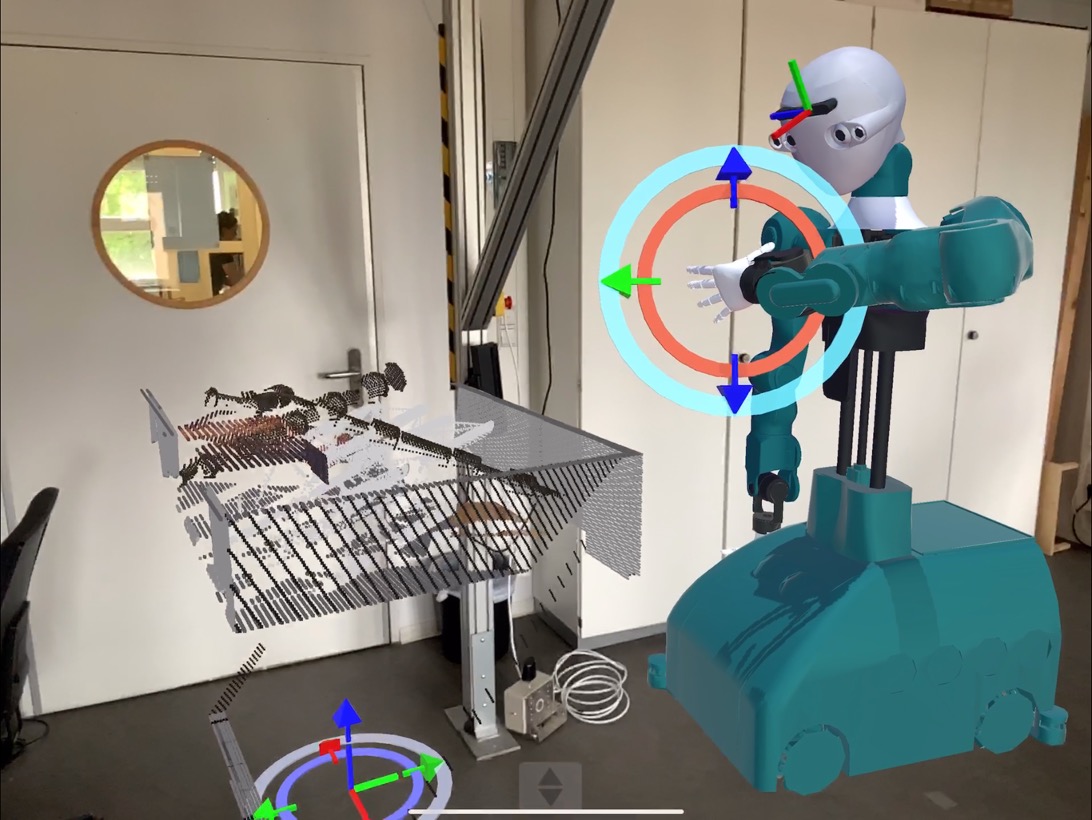}
    \caption{AR grasping simulation with the Armar6 robot \cite{asfour2018armar} from KIT, on an iPad Pro 2018. The top interactive marker represents the target hand position, the bottom marker on the floor is the movable origin of the scene. Left is a simulated point cloud showing the parts to be grasped.}
    \label{fig:markers}
\end{figure*}
\begin{figure*}
    \centering
    \includegraphics[width=0.7\textwidth]{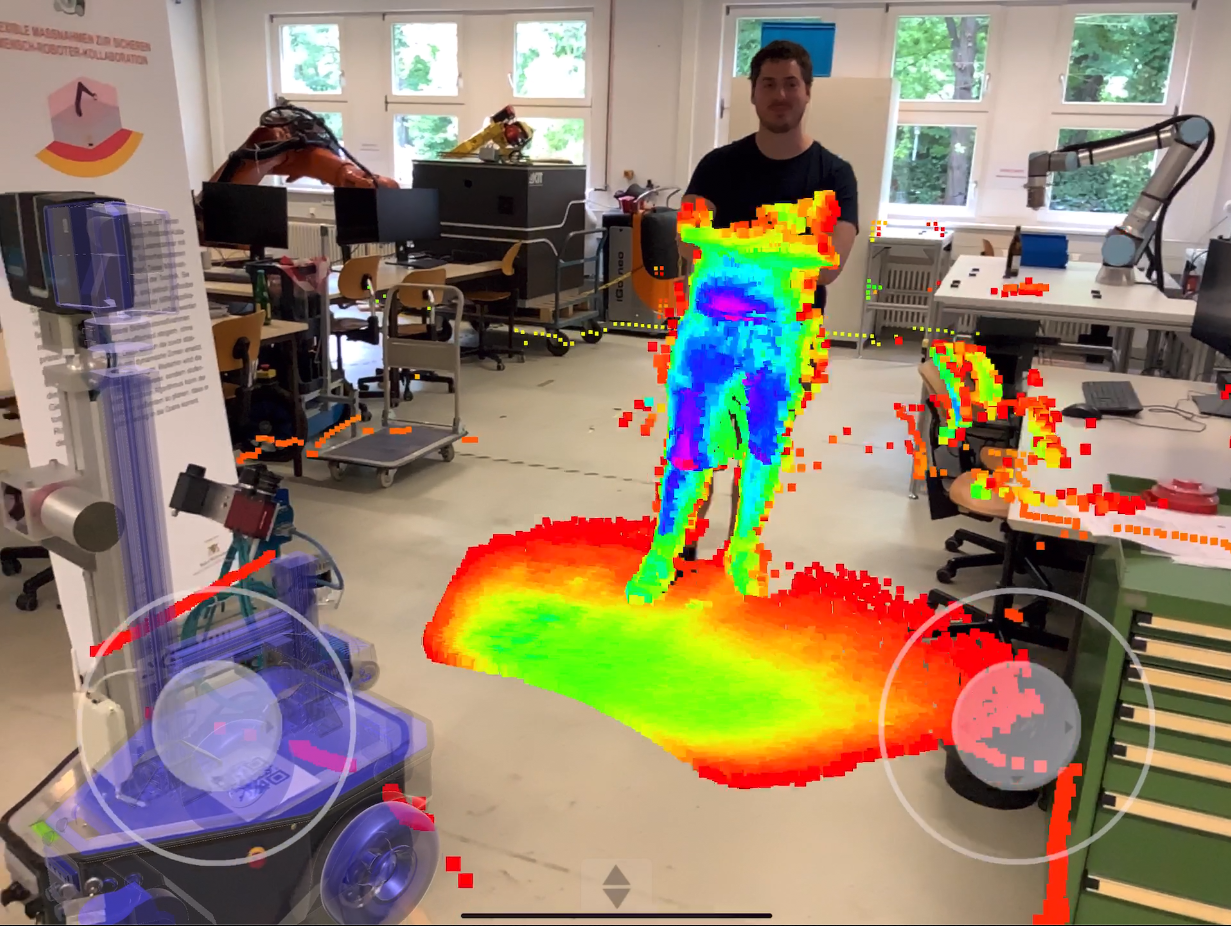}
    \caption{Real AR teleoperation with the Gammabot robot (left, digital twin overlaid in purple) from KIT, on an iPad Pro 2018. The colored dots are a LIDAR pointcloud being overlaid over its real-world measurement sources. The white circles at the sides are virtual joysticks that publish to a \emph{Twist} topic. The marker for registration is visible below the left joystick.}
    \label{fig:teleop}
\end{figure*}

\subsection{Augmented Reality Module}\label{sec:ar}
An important motivation for this project was the ability to visualize ROS data not only in the screen of a mobile device, but also in the real world, around real objects, using AR.
As the Unity engine supports AR natively, this task did not require a significant effort in programming.
However, it is still necessary to find intuitive ways for the user to navigate this visual information without becoming confused.
To achieve this, we focused on two tasks: presentation and environment occlusion.
For presentation, we provide two modalities to view the virtual world.
First, a \emph{scaled view} allows users to present large environments on top of any surface, such as a table, by downscaling the world and anchoring it over a plane.
The scene can then be moved, rotated, or scaled again as necessary (see \Fig{fig:face}).
Second, AR without scaling can be also used by registering (i.e., positioning) the virtual world over the real one, for example when it is necessary to display virtual markers around a real robot.
This registration can be done by hand, e.g., by moving and rotating the virtual scenario until it coincides with the real world, or by using a fiducial marker such as a QR code (as used in \Fig{fig:teleop}).

\begin{figure*}
    \centering
    \includegraphics[width=0.45\textwidth]{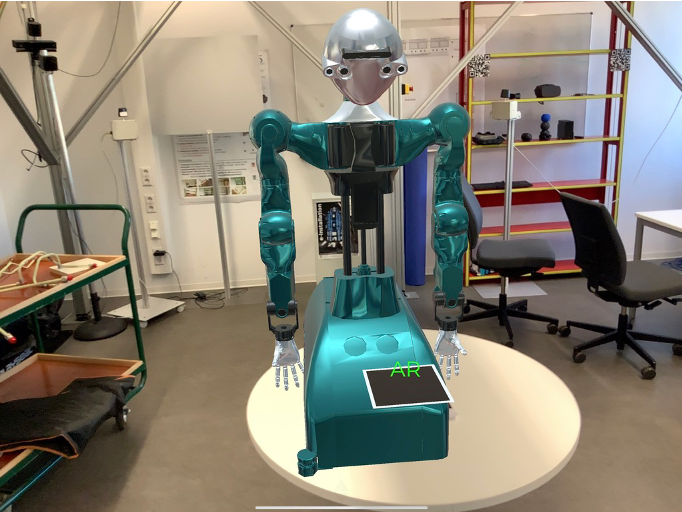}\qquad
    \includegraphics[width=0.45\textwidth]{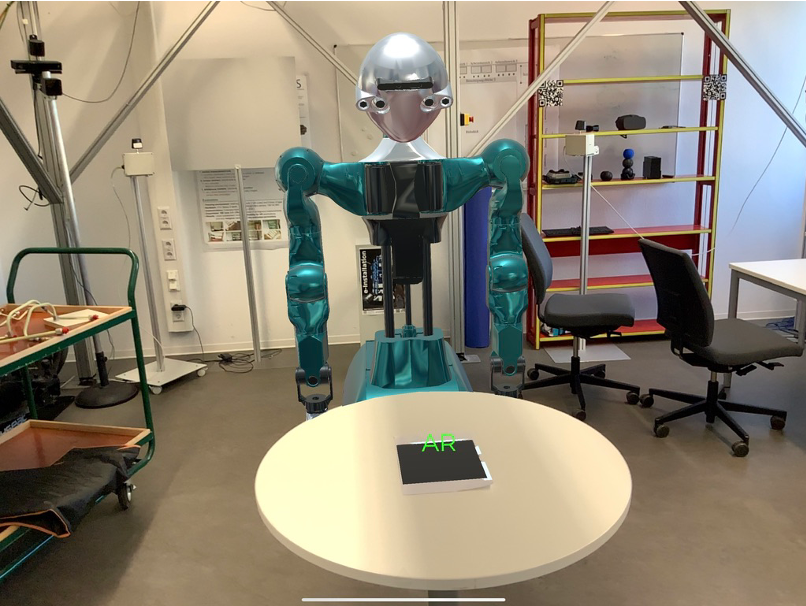}
    \caption{Visualization of a simulated Armar6 behind a real-world table. Because the presentation device cannot provide per-pixel depth information, the robot is rendered on top of the table (left). Using occlusion mode, we can introduce a virtual table marker with the exact size and position as the real-world counterpart. This allows the renderer to display the robot correctly within the scene (right). }
    \label{fig:armer}
\end{figure*}

No matter the presentation, AR can quickly break immersion if occlusion is not handled correctly.
This challenge, which occurs quite often in AR, is visualized in \Fig{fig:armer}, left.
Here, the simulated robot is actually behind the real-world table.
However, because the renderer has no information about depth values in the video feed, the robot appears (quite confusingly) in front of table.
Estimating depth values reliably is a difficult task if the devices have no LIDAR sensors --- i.e., almost every smartphone and tablet.
This, in turn, becomes extremely frustrating for the user, as it is almost impossible to ignore this mismatch with the natural perspective.

To address this, we can exploit the fact that environments where AR visualization is more likely to be used, such as around robots or in laboratories, are generally well known and 3D models of the scene are usually available beforehand.
These assets can be provided to iviz (as markers, robots, or occupancy grid displays), and positioned in the exact same pose as their real-world counterparts
Then, by enabling `occlusion mode' in the displays, instead of drawing them, the renderer only writes values onto the depth buffer, leaving the colors in those pixels untouched.
This allows the renderer to correctly hide any virtual object behind these assets (\Fig{fig:armer}, right), but still allowing for markers in front of them (such as the black marker with the `AR' text).

\section{Conclusions, Challenges, and Future Work}\label{sec:challenges}
In this work, we introduced iviz, a new platform for ROS data visualization aimed at mobile devices.
The main goals were high performance with (relatively) low latency, native AR support, and supporting many display types without depending on intermediary nodes like Rosbridge. 
The project was implemented in Unity, and has already been tested on iOS, Android, UWP, and in desktop operating systems such as Windows, Linux, and macOS.

However, there are some limitations with the concept of a mobile visualization app, especially if we aim to equal the flexibility and generality of rviz.
For example, it is difficult to write `plugins', given that, for security reasons, apps in mobile devices (in particular, those downloaded from app stores) cannot load dynamic code. 
Another important issue is that the visualization code of existing ROS applications cannot be reused in Unity, given that they assume a full ROS installation, a C++/Python programming environment, and libraries such as Qt.
This makes it difficult to operate libraries such as MoveIt \cite{chitta2012moveit} directly in our platform.

Despite the difficulties, there are many merits to having a mobile ROS platform.
Using AR opens up the possibility for new, more intuitive forms of interaction with robots and the environment.
Furthermore, the Unity environment allows it to be easily ported to VR applications, which is the next step in our future work.
For this reason, we believe that iviz can be a great contribution to the ROS community.

\renewcommand*{\bibfont}{\footnotesize}

\DeclareCaseLangs{}

\printbibliography


\end{document}